\documentclass[11pt,twocolumn]{article}
\pdfoutput=1

\usepackage[utf8]{inputenc}
\usepackage[dvipsnames]{xcolor}
\usepackage{hyperref}
\usepackage{acl2017}
\usepackage{times}
\usepackage{amssymb}
\usepackage{latexsym}
\usepackage{comment}
\usepackage{amsmath}
\usepackage{wrapfig}
\usepackage{bm}
\usepackage{graphicx}
\usepackage{booktabs}
\usepackage{todonotes}
\usepackage{multirow}
\usepackage{tablefootnote}
\usepackage{esvect}
\usepackage{cleveref}
\crefformat{section}{\S#2#1#3}

\newcommand\given[1][]{\:#1\vert\:}

\newcommand{\white}[1]{\textcolor{White}{#1}}

\newcommand{\red}[1]{\textcolor{Red}{#1}}
\newcommand{\redbf}[1]{\textbf{\textcolor{Red}{#1}}}

\newcommand{\bluebf}[1]{\textbf{\textcolor{Blue}{#1}}}
\newcommand{\green}[1]{\textcolor{OliveGreen}{#1}}
\newcommand{\greenbf}[1]{\textbf{\textcolor{OliveGreen}{#1}}}

\aclfinalcopy 
\def\aclpaperid{***} 

\title{Incorporating Global Visual Features into\\Attention-Based Neural Machine Translation}

\author{Iacer Calixto\\
	    ADAPT Centre\\
	    Dublin City University\\
	    Glasnevin, Dublin 9\\
	    {\tt\small iacer.calixto@adaptcentre.ie}
	  \And
	Qun Liu\\
	ADAPT Centre\\
  	Dublin City University\\
  	Glasnevin, Dublin 9\\
  \And
	Nick Campbell\\
	ADAPT Centre\\
  	Trinity College Dublin\\
  	College Green, Dublin 2\\
  }

\date{}

\begin{document}

\maketitle

\begin{abstract}
 We introduce multi-modal, attention-based neural machine translation (NMT) models
 which incorporate visual features into different parts of both the encoder and the decoder.
 We utilise global image features extracted using a pre-trained convolutional neural network and incorporate them \textit{(i)} as words in the source sentence, \textit{(ii)} to initialise the encoder hidden state, and \textit{(iii)} as additional data to initialise the decoder hidden state.
 In our experiments, we evaluate how these different strategies to incorporate global image features compare and which ones perform best.
 We also study the impact that adding synthetic multi-modal, multilingual data brings and find that the additional data have a positive impact on multi-modal models.
 We report new state-of-the-art results and our best models also significantly improve on a comparable phrase-based Statistical MT (PBSMT) model trained on the Multi30k data set according to all metrics evaluated.
 To the best of our knowledge, it is the first time a purely neural model significantly improves over a PBSMT model on all metrics evaluated on this data set.
\end{abstract}

\setlength{\belowdisplayskip}{1.0pt} \setlength{\belowdisplayshortskip}{1.0pt}
\setlength{\abovedisplayskip}{1.0pt} \setlength{\abovedisplayshortskip}{1.0pt}
\setlength{\intextsep}{1.5pt}

\section{Introduction}
\label{sec:intro}

Neural Machine Translation (NMT) has recently been proposed as an instantiation of the \emph{sequence to sequence} (\emph{seq2seq}) learning problem~\cite{KalchbrennerBlunsom2013,Choetal2014,SutskeverVinyalsLe2014}.
In this problem, each training example consists of one source and one target variable-length sequence, with no prior information regarding the alignments between the two.
A model is trained to \emph{translate} sequences in the source language into corresponding sequences in the target.
This framework has been successfully used in many different tasks, such as
handwritten text generation~\cite{Graves2013},
image description generation~\cite{Hodoshetal2013,Kirosetal2014b,Maoetal2014,Elliottetal2015,KarpathyFeiFei2015,Vinyalsetal2014},
machine translation~\cite{Choetal2014,SutskeverVinyalsLe2014}
and
video description generation~\cite{Donahueetal2015,Venugopalanetal2015}.

Recently, there has been an increase in the number of natural language generation models that explicitly use \emph{attention-based decoders}, i.e. decoders that model an \emph{intra-sequential} mapping between source and target representations.
For instance,~\newcite{Xuetal2015} proposed an attention-based model
for the task of image description generation where the model learns to \emph{attend to}
specific parts of an image (the source)
as it generates its description (the target).
In MT,
one can intuitively interpret this attention mechanism as inducing
an \emph{alignment} between source and target sentences,
as first proposed by~\newcite{BahdanauChoBengio2015}.
The common idea is to explicitly frame a learning task in which
the decoder learns to attend to the relevant parts of the source sequence
when generating each part of the target sequence.

We are inspired by recent successes in using attention-based models in both image description generation and NMT.
Our main goal in this work is to propose end-to-end multi-modal NMT models which effectively incorporate visual features in different parts of the attention-based NMT framework.
The main contributions of our work are:
\begin{itemize}
 \item We propose novel attention-based multi-modal NMT models which incorporate visual features into the encoder and the decoder.
 
 \item We discuss the impact that adding synthetic multi-modal and multilingual data brings to multi-modal NMT.
 
 \item We show that images bring useful information to an NMT model and report state-of-the-art results.
\end{itemize}

One additional contribution of our work is that
we corroborate previous findings
by~\newcite{Vinyalsetal2014} that suggested that
using image features directly as additional context to update the hidden state of the decoder
(at each time step) leads to overfitting, ultimately preventing learning.

The remainder of this paper is structured as follows.
In~\cref{sec:background} we briefly discuss relevant previous related work.
We then revise the attention-based NMT framework and further expand it into different multi-modal NMT models (\cref{sec:attention-based-neural-machine-translation}).
In~\cref{sec:dataset} we introduce the data sets we use in our experiments. In~\cref{sec:experiments} we detail the hyperparameters, parameter initialisation and other relevant details of our models.
Finally, in~\cref{sec:conclusion} we draw conclusions and provide some avenues for future work.

\subsection{Related work}
\label{sec:background}

Attention-based encoder-decoder models for MT have been actively investigated in recent years.
Some researchers have studied how to improve attention mechanisms~\cite{Luongetal2015,Tuetal2016} and how to train attention-based models to translate between many languages~\cite{Dongetal2015,Firatetal2016}.

There has been some previous related work on using images in tasks involving multilingual and multi-modal natural language generation.
\newcite{Calixtoetal2012} studied how the visual context of a textual description can be helpful in the disambiguation of Statistical MT (SMT) systems.
\newcite{Hitschleretal2016} used image features for re-ranking translations of image descriptions generated by an SMT model and reported significant improvements.
\newcite{Elliottetal2015} generated multilingual descriptions of images
by learning and transferring features between
two independent, non-attentive neural image description models.
\newcite{Luongetal2016} proposed a multi-task learning approach and incorporated neural image description as an auxiliary task to sequence-to-sequence NMT and improved translations in the main translation task.

Multi-modal MT has recently been addressed by the MT community in the form of a shared task~\cite{Speciaetal2016}.
We note that in the official results of this first shared task no submissions based on a purely neural architecture could improve on the phrase-based SMT (PBSMT) baseline.
Nevertheless, researchers have proposed to include global visual features in re-ranking $n$-best lists generated by a PBSMT system or directly in a purely NMT framework with some success~\cite{Caglayanetal2016,CalixtoElliottFrank2016,Libovickyetal2016,Shahetal2016}.
The best results achieved by a purely NMT model in this shared task are those of \newcite{Huangetal2016}, who proposed to use global and regional image features extracted with the VGG19 network.

Similarly to one model we propose,\footnote{This idea has been developed independently by both research groups.} they extract global features for an image, project these features into the vector space of the source words and then add it as a word in the input sequence.
Their best model improves over a strong NMT baseline and is comparable to results obtained with a PBSMT model trained on the same data.
For that reason, their models are used as baselines in our experiments.
Next, we point out some key differences between their models and ours.

\paragraph{Architecture}
Their implementation is based on the attention-based model of~\newcite{Luongetal2015}, which has some differences to that of~\newcite{BahdanauChoBengio2015}, used in our work (\cref{sec:text-only-attention-based-neural-machine-translation}).
Their encoder is a single-layer unidirectional LSTM and they use the last hidden state of the encoder to initialise the decoder's hidden state, therefore indirectly using the image features to do so.
We use a bi-directional recurrent neural network (RNN) with GRU~\cite{Choetal2014b} as our encoder, better encoding the semantics of the source sentence.

\paragraph{Image features}
We include image features separately either as a word in the source sentence~(\cref{sec:image-as-word}) or \emph{directly} for encoder~(\cref{sec:image-encoder-init}) or decoder initialisation (\cref{sec:image-decoder-init}), whereas \newcite{Huangetal2016} only use it as a word.
We also
show it is better to include an image exclusively for the encoder \emph{or} the decoder initialisation (Tables~\ref{tbl:evaluation-translational-flickr30k} and~\ref{tbl:evaluation-backtranslated-flickr30k}).

\paragraph{Data}
\newcite{Huangetal2016} use object detections obtained with the RCNN of \newcite{Girshicketal2014} as additional data, whereas
we study the impact that additional back-translated data brings.

\paragraph{Performance}
All our models outperform \newcite{Huangetal2016}'s according to all metrics evaluated, even when they use additional object detections.
If we use additional back-translated data, the difference becomes even larger.

\section{Attention-based NMT}
\label{sec:attention-based-neural-machine-translation}

In this section, we briefly revise the attention-based NMT framework (\cref{sec:text-only-attention-based-neural-machine-translation}) and expand it into a multi-modal NMT framework (\cref{sec:multimodal-neural-machine-translation}).

\subsection{Text-only attention-based NMT}
\label{sec:text-only-attention-based-neural-machine-translation}

We follow the notation of~\newcite{BahdanauChoBengio2015}
and~\newcite{Firatetal2016} throughout this section.
Given a source sequence ${X = (x_1, x_2, \cdots, x_N)}$ and
its translation ${Y = (y_1, y_2, \cdots, y_M)}$,
an NMT model aims at building a single neural network that translates $X$ into $Y$
by directly learning to model $p(Y \given X)$.
Each $x_i$ is a row index in a source lookup matrix
$\bm{W}_x \in \mathbb{R}^{|V_x| \times d_x}$
(the \emph{source word embeddings matrix})
and each $y_j$ is an index in a target lookup matrix
$\bm{W}_y \in \mathbb{R}^{|V_y| \times d_y}$
(the \emph{target word embeddings matrix}).
$V_x$ and $V_y$ are source and target vocabularies
and $d_x$ and $d_y$ are source and target word embeddings dimensionalities,
respectively.

A bidirectional RNN with GRU is used as the encoder.
A forward RNN $\overrightarrow{\Phi}_{\text{enc}}$
reads $X$ word by word, from left to right,
and generates a sequence of \emph{forward annotation vectors}
${(\overrightarrow{\bm{h}}_1, \overrightarrow{\bm{h}}_2, \cdots, \overrightarrow{\bm{h}}_N)}$
at each encoder time step ${i \in [1,N]}$.
Similarly, a backward RNN $\overleftarrow{\Phi}_{\text{enc}}$
reads $X$
from right to left,
word by word, and generates a sequence of \emph{backward annotation vectors}
${(\overleftarrow{\bm{h}}_1, \overleftarrow{\bm{h}}_2, \cdots, \overleftarrow{\bm{h}}_N)}$, as in~(\ref{eq:bidirectional-rnn}):
\begin{align}
\label{eq:bidirectional-rnn}
    \overrightarrow{\bm{h}_i} = \overrightarrow{\Phi}_{\text{enc}}
    \big( \bm{W}_x[x_i], \overrightarrow{\bm{h}}_{i-1} \big), \notag\\
    \overleftarrow{\bm{h}_i} = \overleftarrow{\Phi}_{\text{enc}}
    \big( \bm{W}_x[x_i], \overleftarrow{\bm{h}}_{i+1} \big).
\end{align}
The final annotation vector for a given time step $i$ is the concatenation of forward and backward vectors $\bm{h}_i = \big[ \overrightarrow{\bm{h}_i}; \overleftarrow{\bm{h}_i} \big]$.

In other words, each source sequence $X$ is encoded into a sequence of annotation vectors ${h = (\bm{h}_1, \bm{h}_2, \cdots, \bm{h}_N)}$, which are in turn used by the decoder: essentially a neural language model (LM)~\cite{Bengioetal2003} conditioned on the previously emitted words and the source sentence via an attention mechanism.

At each time step $t$ of the decoder, we compute a \emph{time-dependent} context vector $\bm{c}_t$ based on the annotation vectors $h$, the decoder's previous hidden state $\bm{s}_{t-1}$ and the target word $\tilde{y}_{t-1}$ emitted by the decoder in the previous time step.\footnote{At training time, the correct previous target word $y_{t-1}$ is known and therefore used instead of $\tilde{y}_{t-1}$.
At test or inference time, $y_{t-1}$ is not known and $\tilde{y}_{t-1}$ is used instead.
\newcite{Bengioetal2015} discussed problems that may arise from this difference between training and inference distributions.
}

We follow~\newcite{BahdanauChoBengio2015}
and use a single-layer feed-forward network
to compute an \emph{expected alignment} $\bm{e}_{t,i}$
between each source annotation vector $\bm{h}_i$
and the target word to be emitted at the current time step $t$, as in~(\ref{eq:expected-alignment}):
\begin{equation}
\label{eq:expected-alignment}
  \bm{e}_{t,i} = {\bm{v}_a}^T \tanh( \bm{U}_a \bm{s}_{t-1} + \bm{W}_a \bm{h}_i).
\end{equation}

\begin{figure}[t!]
 \centering
 \includegraphics[width=0.35\textwidth]{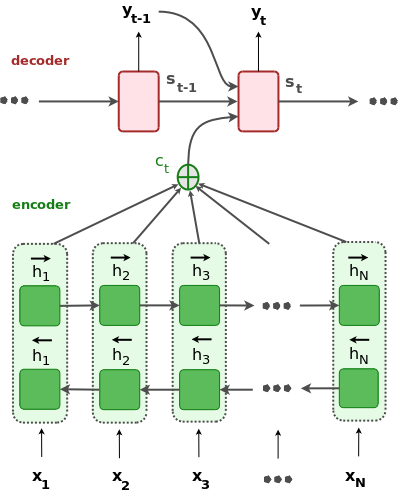}
 \caption{Computation of the decoder's hidden state $\bm{s}_t$
 using the attention mechanism.}
 \label{fig:attention-mechanism}
\end{figure}

In Equation~(\ref{eq:normalised-expected-alignment}),
these expected alignments are further normalised and converted into
probabilities:
\begin{equation}
\label{eq:normalised-expected-alignment}
  \bm{\alpha}_{t,i} = \frac{\exp{(\bm{e}_{t,i})}}{ \sum_{j=1}^{N}{\exp{(\bm{e}_{t,j})}} },
\end{equation}

\noindent
where $\alpha_{t,i}$ are called the model's \emph{attention weights},
which are in turn used in computing the time-dependent context vector
${\bm{c}_t = \sum_{i=1}^{N}{ \bm{\alpha}_{t,i} \bm{h}_i }}$.
Finally, the context vector $\bm{c}_t$ is used in computing the decoder's hidden state $\bm{s}_t$ for the current time step $t$, as shown in Equation~(\ref{eq:decoder}):
\begin{equation}
\label{eq:decoder}
    \bm{s}_t = \Phi_{\text{dec}}( \bm{s}_{t-1}, \bm{W_y}[\tilde{y}_{t-1}], \bm{c}_t ),
\end{equation}

\noindent
where $\bm{s}_{t-1}$ is the decoder's previous hidden state,
$\bm{W_y}[\tilde{y}_{t-1}]$ is the embedding of the word emitted in the previous time step,
and $\bm{c}_t$ is the updated time-dependent context vector.
In Figure~\ref{fig:attention-mechanism}
we illustrate the computation of
the decoder's hidden state $\bm{s}_t$.

We use a single-layer feed-forward neural network to initialise the decoder's hidden state $\bm{s}_0$ at time step $t=0$ and feed it the concatenation of the last hidden states of the encoder's forward RNN  ($\overrightarrow{\Phi}_{\text{enc}}$) and backward RNN ($\overleftarrow{\Phi}_{\text{enc}}$), as in~(\ref{eq:decoder-initial-state}):
\begin{equation}
\label{eq:decoder-initial-state}
\bm{s}_0 = \tanh\big( \bm{W}_{di} [\overleftarrow{\bm{h}}_1; \overrightarrow{\bm{h}_N}] + \bm{b}_{di} \big),
\end{equation}
\noindent
where $\bm{W}_{di}$ and $\bm{b}_{di}$ are model parameters.
Since RNNs normally better store information about recent inputs in comparison to more distant ones~\cite{HochreiterSchmidhuber1997,BahdanauChoBengio2015}, we expect to initialise the decoder's hidden state with a strong source sentence representation, i.e. a representation with a strong focus on both the first and the last tokens in the source sentence.

\subsection{Multi-modal NMT (MNMT)}
\label{sec:multimodal-neural-machine-translation}

Our models can be seen as
expansions of the attention-based NMT framework described in~\cref{sec:attention-based-neural-machine-translation} with
the addition of a \emph{visual component} to incorporate image features.

Simonyan and Zisserman~\shortcite{SimonyanZisserman2014} trained and evaluated an extensive set of deep convolutional neural network (CNN) models for classifying images into one out of the $1000$ classes in ImageNet~\cite{Russakovskyetal2014}.
We use their 19-layer VGG network (VGG$19$) to extract image feature vectors for all images in our dataset.
We feed an image to the pre-trained VGG$19$ network and use the $4096$D activations of the penultimate fully-connected layer
FC$7$\footnote{We use the activations of the FC$7$ layer, which encode information about the entire image, of the VGG$19$ network (configuration E) in~\newcite{SimonyanZisserman2014}'s paper.}
as our \emph{image feature vector}, henceforth referred to as $\bm{q}$.

We propose three different methods to incorporate images into the attentive NMT framework:
using an image as words in the source sentence (\Cref{sec:image-as-word}), using an image to initialise the source language encoder (\cref{sec:image-encoder-init}) and the target language decoder (\cref{sec:image-decoder-init}).

We also evaluated a fourth mechanism to incorporate images into NMT, namely to use an image as one of the different contexts available to the decoder at each time step of the decoding process.
We add the image features directly as an additional context,
in addition to $\bm{W}_{y}[\tilde{y}_{t-1}]$, $\bm{s}_{t-1}$ and $\bm{c}_t$,
to compute the hidden state $\bm{s}_t$ of the decoder at a given time step $t$.
We corroborate previous findings by~\newcite{Vinyalsetal2014} in that adding the image features as such causes the model to overfit, ultimately preventing learning.\footnote{For comparison, translations for the translated Multi30k test set (described in \cref{sec:dataset}) achieve just 3.8 BLEU~\cite{Papinenietal2002}, 15.5 METEOR~\cite{DenkowskiLavie2014} and 93.0 TER~\cite{Snoveretal2006}.}

\subsubsection{Images as source words: \texorpdfstring{IMG$_{\text{W}}$}{}}
\label{sec:image-as-word}

One way we propose to incorporate images into the encoder is to project an image feature vector into the space of the words of the source sentence.
We use the projected image as the first and/or last word of the source sentence and let the attention model learn when to attend to the image representation.
Specifically, given the global image feature vector $\bm{q} \in \mathbb{R}^{4096}$,
we compute~(\ref{eq:image-projection}):
\begin{align}\label{eq:image-projection}
  \bm{d} = \bm{W}_I^2 \cdot ( \bm{W}_I^1 \cdot \bm{q} + \bm{b}_I^1 ) + \bm{b}_I^2 ,
\end{align}
\noindent
where $\bm{W}_I^1 \in \mathbb{R}^{4096 \times 4096}$ and $\bm{W}_I^2 \in \mathbb{R}^{4096 \times d_x}$ are image transformation matrices, $\bm{b}_I^1 \in \mathbb{R}^{4096}$ and $\bm{b}_I^2 \in \mathbb{R}^{d_x}$ are bias vectors, and $d_x$ is the source words vector space dimensionality, all trained with the model.
We then directly use $\bm{d}$ as words in the source words vector space:
as the first word only (model IMG$_{1\text{W}}$), and
as the first and last words of the source sentence (model IMG$_{2\text{W}}$).

\begin{figure}[t!]
 \centering
 \includegraphics[width=0.32\textwidth]{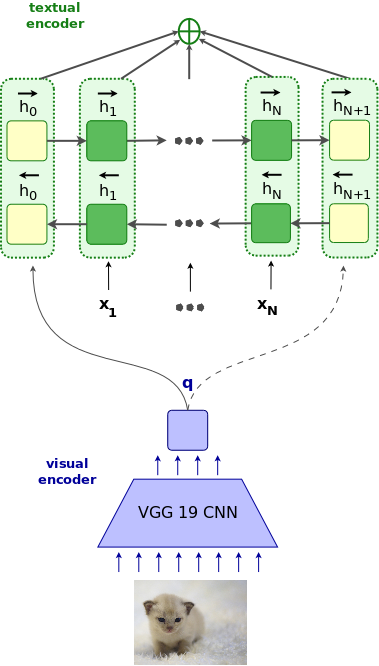}
 \caption{An encoder bidirectional RNN that uses image features
 as words in the source sequence.}
 \label{fig:image-as-word}
\end{figure}

An illustration of this idea is given in Figure~\ref{fig:image-as-word}, where a source sentence that originally contained $N$ tokens, after including the image as source words will contain
$N+1$ tokens (model IMG$_{1\text{W}}$) or
$N+2$ tokens (model IMG$_{2\text{W}}$).
In model IMG$_{1\text{W}}$, the image is projected as the first source word only (solid line in Figure~\ref{fig:image-as-word});
 in model IMG$_{2\text{W}}$, it is projected into the source words space as both first and last words (both solid and dashed lines in Figure~\ref{fig:image-as-word}).
 
Given a source sequence $X = (x_1, x_2, \cdots, x_N)$,
we concatenate the transformed image vector $\bm{d}$ to $\bm{W}_x[X]$ and apply the forward and backward encoder RNN passes, generating hidden vectors as in Figure~\ref{fig:image-as-word}.
When computing the context vector $\bm{c}_t$ (Equations~(\ref{eq:expected-alignment}) and~(\ref{eq:normalised-expected-alignment})), we effectively make use of the transformed image vector, i.e. the $\bm{\alpha}_{t,i}$ attention weight parameters will use this information to attend or not to the image features.

By including images into the encoder in models
IMG$_{1\text{W}}$ and IMG$_{2\text{W}}$, our intuition is that
\textit{(i)} by including the image as the \emph{first word}, we propagate image features into the source sentence vector representations when applying the forward RNN $\overrightarrow{\Phi}_{\text{enc}}$ (vectors $\overrightarrow{\bm{h}_i}$), and
\textit{(ii)} by including the image as the \emph{last word}, we propagate image features into the source sentence vector representations when applying the backward RNN $\overleftarrow{\Phi}_{\text{enc}}$ (vectors $\overleftarrow{\bm{h}_i}$).

\subsubsection{Images for encoder initialisation: \texorpdfstring{IMG$_{\text{E}}$}{}}
\label{sec:image-encoder-init}

\begin{figure}[t!]
 \centering
 \includegraphics[width=0.4\textwidth]{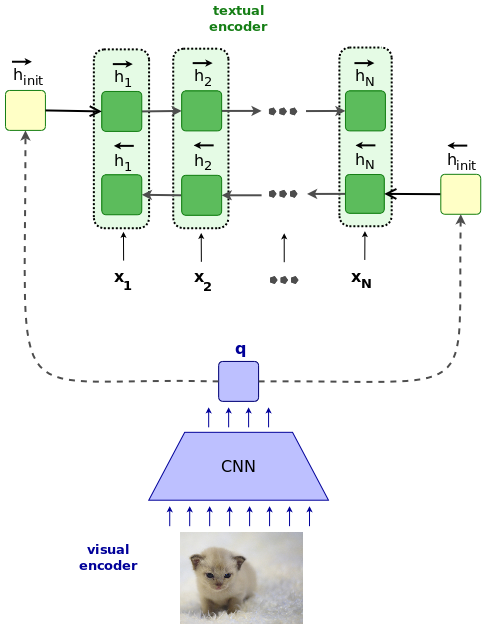}
 \caption{Using an image to initialise the encoder hidden states.}
 \label{fig:image-encoder-init}
\end{figure}

In the original attention-based NMT model described in~\cref{sec:attention-based-neural-machine-translation}, the hidden state of the encoder is initialised with the zero vector $\vv{0}$.
Instead, we propose to use two new single-layer feed-forward neural networks to compute the initial states of the forward RNN $\overrightarrow{\Phi}_{\text{enc}}$ and the backward RNN $\overleftarrow{\Phi}_{\text{enc}}$, respectively, as illustrated in Figure~\ref{fig:image-encoder-init}.

Similarly to~\cref{sec:image-as-word}, given a global image feature vector $\bm{q} \in \mathbb{R}^{4096}$, we compute a vector $\bm{d}$ using Equation~(\ref{eq:image-projection}), only this time the parameters $\bm{W}_I^2$ and $\bm{b}_I^2$ project the image features into the same dimensionality as the textual encoder hidden states.

The feed-forward networks used to initialise the encoder hidden state are computed as in (\ref{eq:encoder-initial-state-multimodal}):
\begin{align}
\label{eq:encoder-initial-state-multimodal}
\overleftarrow{\bm{h}}_\text{init} = \tanh\big( \bm{W}_{f}\bm{d} + \bm{b}_{f} \big),\notag\\
\overrightarrow{\bm{h}}_\text{init} = \tanh\big( \bm{W}_{b}\bm{d} + \bm{b}_{b} \big),
\end{align}
\noindent
where $\bm{W}_{f}$ and $\bm{W}_{b}$ are multi-modal projection matrices that project the image features $\bm{d}$ into the encoder forward and backward hidden states dimensionality, respectively, and $\bm{b}_{f}$ and $\bm{b}_{b}$ are bias vectors.

\subsubsection{Images for decoder initialisation: \texorpdfstring{IMG$_{\text{D}}$}{}}
\label{sec:image-decoder-init}

To incorporate an image into the decoder, we introduce a new single-layer feed-forward neural network to be used instead of the one described in Equation~\ref{eq:decoder-initial-state}.
Originally, the decoder's initial hidden state was computed using the concatenation of the last hidden states of the encoder forward RNN  ($\overrightarrow{\Phi}_{\text{enc}}$) and backward RNN ($\overleftarrow{\Phi}_{\text{enc}}$), respectively $\overrightarrow{\bm{h}_N}$ and $\overleftarrow{\bm{h}}_1$.

\begin{figure}[t!]
 \centering
 \includegraphics[width=0.4\textwidth]{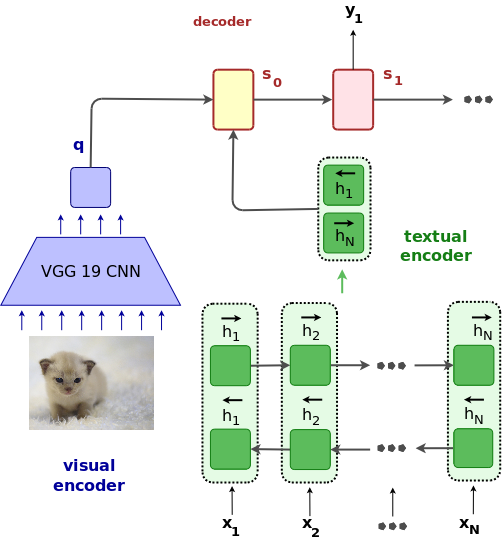}
 \caption{Image as additional data to initialise the decoder
 hidden state $s_0$.}
 \label{fig:image-decoder-init}
\end{figure}

Our proposal is that we include the image features as additional input to initialise the decoder hidden state at time step $t=0$, as in~(\ref{eq:decoder-initial-state-multimodal}):
\begin{equation}
\label{eq:decoder-initial-state-multimodal}
\bm{s}_0 = \tanh\big( \bm{W}_{di} [\overleftarrow{\bm{h}}_1; \overrightarrow{\bm{h}_N}] + \bm{W}_{m}\bm{d} + \bm{b}_{di} \big),
\end{equation}
\noindent
where $\bm{W}_{m}$ is a multi-modal projection matrix that projects the image features $\bm{d}$ into the decoder hidden state dimensionality and $\bm{W}_{di}$ and $\bm{b}_{di}$ are the same as in Equation~(\ref{eq:decoder-initial-state}).

Once again we compute $\bm{d}$ by applying Equation~(\ref{eq:image-projection}) onto a global image feature vector \mbox{$\bm{q} \in \mathbb{R}^{4096}$},
only this time the parameters $\bm{W}_I^2$ and $\bm{b}_I^2$ project the image features into the same dimensionality as the decoder hidden states.
We illustrate this idea in Figure~\ref{fig:image-decoder-init}.

\section{Data set}
\label{sec:dataset}

Our multi-modal NMT models need bilingual sentences accompanied by one or more images as training data.
The original Flickr30k data set contains $30$k images and $5$ English sentence descriptions for each image~\cite{Youngetal2014}.
We use the translated and the comparable Multi30k datasets \cite{ElliottFrankSimaanSpecia2016}, henceforth referred to as M30k$_\text{T}$ and M30k$_\text{C}$, respectively, which are multilingual expansions of the original Flickr30k.

For each of the 30k images in the Flickr30k, the M30k$_\text{T}$ has one of its English descriptions manually translated into German by a professional translator. Training, validation and test sets contain 29k, 1014 and 1k images, respectively, each accompanied by one sentence pair (the original English sentence and its German translation).
For each of the 30k images in the Flickr30k, the M30k$_\text{C}$ has five descriptions in German collected independently of the English descriptions. Training, validation and test sets contain 29k, 1014 and 1k images, respectively, each accompanied by five sentences in English and five sentences in German.

We use the scripts in the Moses SMT Toolkit~\cite{Koehnetal2007} to normalise, truecase and tokenize English and German descriptions
and we also convert space-separated tokens into subwords~\cite{Sennrichetal2016}.
All models use a common vocabulary of 83,093 English and 91,141 German subword tokens.
If sentences in English or German are longer than 80 tokens, they are discarded.

We use the entire M30k$_\text{T}$
training set for training, its validation set for model selection with BLEU, and its test set to evaluate our models.
In order to study the impact that additional training data brings to the models,
we use the baseline model described in~\cref{sec:attention-based-neural-machine-translation} trained on the textual part of the M30k$_\text{T}$ data set (German$\rightarrow$English) without the images to build a back-translation model~\cite{Sennrichetal2016a}.
We back-translate the $145$k German descriptions in the M30k$_\text{C}$ into English and include the triples (synthetic English description, German description, image) as additional training data.

We train models to translate from English into German and report evaluation of cased, tokenized sentences with punctuation.

\section{Experimental setup}
\label{sec:experiments}

Our encoder is a bidirectional RNN with GRU
(one 1024D single-layer forward RNN and one 1024D single-layer backward RNN).
Source and target word embeddings are 620D each
and both are trained jointly with our model.
All non-recurrent matrices are initialised by sampling from
a Gaussian distribution $(\mu=0, \sigma=0.01)$,
recurrent matrices are orthogonal and
bias vectors are all initialised to zero.
Our decoder RNN also uses GRU and is a neural LM~\cite{Bengioetal2003} conditioned on its previous emissions and the source sentence by means of the source attention mechanism.

Image features are obtained by feeding images to the pre-trained VGG$19$ network of Simonyan and Zisserman~\shortcite{SimonyanZisserman2014} and using the activations of the penultimate fully-connected layer FC$7$.
We apply dropout with a probability of $0.2$ in both source and target word embeddings and with a probability of $0.5$ in the image features (in all MNMT models), in the encoder and decoder RNNs inputs and recurrent connections, and before the readout operation in the decoder RNN.
We follow~\newcite{Gal2015} and apply dropout to the encoder bidirectional RNN and decoder RNN using the same mask in all time steps.

Our models are trained using
stochastic gradient descent with Adadelta~\cite{Zeiler2012} and minibatches of size 40, where each training instance consists of one English sentence, one German sentence and one image.
We apply early stopping for model selection based on BLEU scores, so that if a model does not improve on BLEU in the validation set for more than 20 epochs, training is halted.

We evaluate our models' translation quality quantitatively in terms of BLEU$4$~\cite{Papinenietal2002}, METEOR~\cite{DenkowskiLavie2014}, TER~\cite{Snoveretal2006}, and chrF3 scores\footnote{We specifically compute character 6-gram F3 scores.}~\cite{Popovic2015} and we report statistical significance for the three first metrics using approximate randomisation computed with \mbox{MultEval}~\cite{Clarketal2011}.

As our main baseline we train an attention-based NMT model (\cref{sec:attention-based-neural-machine-translation}) in which only the textual part of M30k$_\text{T}$ is used for training.
We also train a PBSMT model built with Moses on the same data.
The LM is a 5--gram LM with modified Kneser-Ney smoothing~\cite{KneserNey1995} trained on the German side of the  M30k$_\text{T}$ dataset.
We use minimum error rate training~\cite{Och2003} for tuning the model parameters for BLEU scores.
Our third baseline is the best comparable multi-modal model by \newcite{Huangetal2016} and also their best model with additional object detections: respectively models \texttt{m1} (image at head) and \texttt{m3} in the authors' paper.

\subsection{Results}
\label{sec:results}

\begin{table}[t!]
  \centering
  \resizebox{\linewidth}{!} {
  \begin{tabular}{lllll}
   \toprule
   & BLEU$4$$\uparrow$ & METEOR$\uparrow$ & TER$\downarrow$ & chrF3$\uparrow$ \\
   \toprule
   PBSMT &
   32.9 &
   \underline{54.1} &
   \underline{45.1} &
   \underline{67.4} \\
   
   NMT &
   \underline{33.7} &
   52.3 &
   46.7 &
   64.5 \\
   
   \midrule
   
   Huang
   & 35.1 & 52.2 & --- & --- \\
   
   + RCNN
   & 36.5 & 54.1 & --- & --- \\
   
   \midrule
   
   IMG$_{1\text{W}}$ &
   37.1$^\dagger$$^\ddagger$ \small\greenbf{($\uparrow$ 3.4)} &
   54.5\white{$^\dagger$}$^\ddagger$ \small\greenbf{($\uparrow$ 0.4)} &
   42.7$^\dagger$$^\ddagger$ \small\greenbf{($\downarrow$ 2.4)} &
   66.9 \small\redbf{($\downarrow$ 0.5)}\\
   
   IMG$_{2\text{W}}$ &
   36.9$^\dagger$$^\ddagger$ \small\greenbf{($\uparrow$ 3.2)} &
   54.3\white{$^\dagger$}$^\ddagger$ \small\greenbf{($\uparrow$ 0.2)} &
   \textbf{41.9}$^\dagger$$^\ddagger$ \small\greenbf{($\downarrow$ 3.2)} &
   66.8 \small\redbf{($\downarrow$ 0.6)} \\
   
   IMG$_{\text{E}}$ &
   37.1$^\dagger$$^\ddagger$ \small\greenbf{($\uparrow$ 3.4)} &
   55.0$^\dagger$$^\ddagger$ \small\greenbf{($\uparrow$ 0.9)} &
   43.1$^\dagger$$^\ddagger$ \small\greenbf{($\downarrow$ 2.0)} &
   67.6 \small\greenbf{($\uparrow$ 0.2)}\\
   
   IMG$_{\text{D}}$ &
   \textbf{37.3}$^\dagger$$^\ddagger$ \small\greenbf{($\uparrow$ 3.6)} &
   \textbf{55.1}$^\dagger$$^\ddagger$ \small\greenbf{($\uparrow$ 1.0)} &
   42.8$^\dagger$$^\ddagger$ \small\greenbf{($\downarrow$ 2.3)} &
   \textbf{67.7} \small\greenbf{($\uparrow$ 0.3)}\\
   
   IMG$_{2\text{W+D}}$ &
   35.7$^\dagger$$^\ddagger$ \small\greenbf{($\uparrow$ 2.0)} &
   53.6\white{$^\dagger$}$^\ddagger$ \small\redbf{($\downarrow$ 0.5)} &
   43.3$^\dagger$$^\ddagger$ \small\greenbf{($\downarrow$ 1.8)} &
   66.2 \small\redbf{($\downarrow$ 1.2)}\\
   
   IMG$_{\text{E+D}}$ &
   37.0$^\dagger$$^\ddagger$ \small\greenbf{($\uparrow$ 3.3)} &
   54.7\white{$^\dagger$}$^\ddagger$ \small\greenbf{($\uparrow$ 0.6)} &
   42.6$^\dagger$$^\ddagger$ \small\greenbf{($\downarrow$ 2.5)} &
   67.2 \small\redbf{($\downarrow$ 0.2)}\\
   
   \bottomrule
  \end{tabular}
  }
  \caption{
  BLEU$4$, METEOR, chrF3 (higher is better) and TER scores (lower is better) on the M30k$_\text{T}$
  test set for the two text-only baselines PBSMT and NMT, the two multi-modal NMT models by \newcite{Huangetal2016}
  and our MNMT models that:
  \textit{(i)} use images as words in the source sentence (IMG$_{1\text{W}}$, IMG$_{2\text{W}}$),
  \textit{(ii)} use images to initialise the encoder (IMG$_{\text{E}}$), and
  \textit{(iii)} use images as additional data to initialise the decoder (IMG$_{\text{D}}$).
  Best text-only baselines are underscored and best overall results appear in bold.
  We highlight in parentheses the improvements brought by our models compared to the best corresponding text-only baseline score.
  Results differ significantly from PBSMT baseline ($\dagger$) or NMT baseline ($\ddagger$) with $p=0.05$.
  }
  \label{tbl:evaluation-translational-flickr30k}
\end{table}

The Multi30K dataset contains images and bilingual descriptions.
Overall, it is a small dataset with a small vocabulary whose sentences have simple syntactic structures and not much ambiguity~\cite{ElliottFrankSimaanSpecia2016}.
This is reflected in the fact that even the simplest baselines perform fairly well on it, i.e. the smallest BLEU score of 32.9 is that of the PBSMT model, which is still good for translating into German.

From Table~\ref{tbl:evaluation-translational-flickr30k} we see that our multi-modal models perform well, with models IMG$_\text{E}$ and IMG$_\text{D}$ improving on both baselines according to all metrics analysed.
We also note that all models but IMG$_\text{2W+D}$ perform consistently better than the strong multi-modal NMT baseline of \newcite{Huangetal2016}, even when this model has access to more data (+RCNN features).\footnote{In fact, model IMG$_\text{2W+D}$ still improves on the multi-modal baseline of \newcite{Huangetal2016} when trained on the same data.}
Combining image features in the encoder and the decoder at the same time (last two entries in Table~\ref{tbl:evaluation-translational-flickr30k}) does not seem to improve results compared to using the image features in only the encoder or the decoder.
To the best of our knowledge, it is the first time a purely neural model significantly improves over a PBSMT model in all metrics on this data set.

Arguably, the main downside of applying multi-modal NMT in a real-world scenario is the small amount of publicly available training data ($\sim$$30$k), which restricts its applicability.
For that reason, we back-translated the German sentences in the M30k$_\text{C}$ and created additional $145$k synthetic triples (synthetic English sentence, original German sentence and image).

In Table~\ref{tbl:evaluation-backtranslated-flickr30k}, we present results for some of the models evaluated in Table~\ref{tbl:evaluation-translational-flickr30k} but when also trained on the additional data.
In order to add more data to the PBSMT baseline, we simply added the German sentences in the M30k$_\text{C}$ as additional data to train the LM.\footnote{Adding the synthetic sentence pairs to train the baseline PBSMT model, as we did with all neural MT models, deteriorated the results.}
Both our models IMG$_\text{E}$ and IMG$_\text{D}$ that use global image features to initialise the encoder and the decoder, respectively, improve significantly according to BLEU, METEOR and TER with the additional back-translated data,
and also achieved better chrF3 scores.
Model IMG$_\text{2W}$, that uses images as words in the source sentence, does not significantly differ in BLEU, METEOR or TER ($p=0.05$), but achieves a lower chrF3 score than the comparable PBSMT model.
Although model IMG$_\text{2W}$ trained on only the original data has the best TER score ($=41.9$), both models IMG$_\text{E}$ and IMG$_\text{D}$ perform comparably with the additional back-translated data ($=41.4$ and $41.6$, respectively), though the difference between the latter and the former is still not statistically significant ($p=0.05$).

We see in Tables~\ref{tbl:evaluation-translational-flickr30k} and~\ref{tbl:evaluation-backtranslated-flickr30k} that our models that use images directly to initialise either the encoder or the decoder are the only ones to consistently outperform the PBSMT baseline according to the chrF3 metric, a character-based metric that includes both precision and recall, and has a recall bias.
That is also a noteworthy finding, since chrF3 is the only character-level metric we use, and it has shown a high correlation with human judgements~\cite{Stanojevicetal2015}.

In Table~\ref{tbl:translation-examples} we see translations for two entries in the test M30k set.
In the first entry,
although the reference translation is incorrect---there is just one dog in the image---, the multi-modal models translated it correctly.
In the second entry, the last three multi-modal models extrapolate the reference+image and describe ``ceremony'' as a ``wedding ceremony'' (IMG$_\text{2W}$) and as an ``Olympics ceremony'' (IMG$_\text{E}$ and IMG$_\text{D}$).
This could be due to the fact that the training set is small, depicts a small variation of different scenes and contains different forms of biasses~\cite{Miltenburg2015}.

We note that the idea of using images as words in the source sentence, also entertained by~\newcite{Huangetal2016}, does not perform as well as directly using the images in the encoder or decoder initialisation.
The fact that multi-modal NMT models can benefit from back-translated data
is also an interesting finding.

\begin{table}[t!]
  \centering
  \resizebox{\linewidth}{!} {
  \begin{tabular}{lllll}
   \toprule
   & BLEU$4$$\uparrow$ & METEOR$\uparrow$ & TER$\downarrow$ & chrF3$\uparrow$ \\
   \toprule
   
   \multicolumn{5}{l}{original training data}\\
   \midrule
   
   IMG$_{2\text{W}}$ &
   36.9 & 54.3 & 41.9 & 66.8 \\
   
   IMG$_{\text{E}}$ &
   37.1 & 55.0 & 43.1 & 67.6 \\
   
   IMG$_{\text{D}}$ &
   37.3 & 55.1 & 42.8 & 67.7 \\
   
   \midrule
   \multicolumn{5}{l}{+ back-translated training data}\\
   \midrule
   PBSMT & 34.0             & \underline{55.0} & 44.7 & \underline{68.0} \\
   NMT & \underline{35.5} & 53.4             & \underline{43.3} & 65.3 \\
   \midrule
   
   IMG$_{2\text{W}}$ &
   36.7$^\dagger$$^\ddagger$
   \small\greenbf{($\uparrow$ 1.2)} &
   54.6\white{$^\dagger$}$^\ddagger$
   \small\redbf{($\downarrow$ 0.4)} &
   42.0$^\dagger$$^\ddagger$
   \small\greenbf{($\downarrow$ 1.3)} &
   66.8 \small\redbf{($\downarrow$ 1.2)} \\
   
   IMG$_{\text{E}}$ &
   \textbf{38.5}$^\dagger$$^\ddagger$
   \small\greenbf{($\uparrow$ 3.0)} &
   55.7$^\dagger$$^\ddagger$
   \small\greenbf{($\uparrow$ 0.9)} &
   \textbf{41.4}$^\dagger$$^\ddagger$
   \small\greenbf{($\downarrow$ 1.9)} &
   68.3 \small\greenbf{($\uparrow$ 0.3)} \\
   
   IMG$_{\text{D}}$ &
   \textbf{38.5}$^\dagger$$^\ddagger$
   \small\greenbf{($\uparrow$ 3.0)} &
   \textbf{55.9}$^\dagger$$^\ddagger$
   \small\greenbf{($\uparrow$ 1.1)} &
   41.6$^\dagger$$^\ddagger$
   \small\greenbf{($\downarrow$ 1.7)} &
   \textbf{68.4} \small\greenbf{($\uparrow$ 0.4)} \\
   
   \midrule
   \multicolumn{4}{l}{\textbf{Improvements} (original vs. + back-translated)}\\
   \midrule
   
   IMG$_\text{2W}$&
   {\small\red{$\downarrow$ 0.2}} &
   {\small\green{$\uparrow$ 0.1}} &
   {\small\red{$\uparrow$ 0.1}} &
   {\small\green{$\uparrow$ 0.0}} \\
   
   IMG$_\text{E}$&
   {\small\greenbf{$\uparrow$ 1.4}}&
   {\small\green{$\uparrow$ 0.7}}&
   {\small\greenbf{$\downarrow$ 1.8}}&
   {\small\greenbf{$\uparrow$ 0.7}}\\

   IMG$_\text{D}$&
   {\small\green{$\uparrow$ 1.2}}&
   {\small\greenbf{$\uparrow$ 0.8}}&
   {\small\green{$\downarrow$ 1.2}}&
   {\small\greenbf{$\uparrow$ 0.7}}\\
   
   \bottomrule
  \end{tabular}
  }
  \caption{BLEU$4$, METEOR, TER and chrF3 scores on the M30k$_\text{T}$ test set for models trained on original and additional back-translated data.
  Best text-only baselines are underscored and best overall results in bold.
  We highlight in parentheses the improvements brought by our models compared to the best baseline score.
  Results differ significantly from PBSMT baseline ($\dagger$) or NMT baseline ($\ddagger$) with $p=0.05$.
  We also show the improvements each model yields in each metric when only trained on the original M30k$_\text{T}$ training set vs. also including additional back-translated data.
  }
  \label{tbl:evaluation-backtranslated-flickr30k}
\end{table}

\begin{table}[t!]
  \centering
  \resizebox{\linewidth}{!} {
    \begin{tabular}{ll}
      \toprule
      %
      
      \multicolumn{2}{c}{\includegraphics[height=130pt]{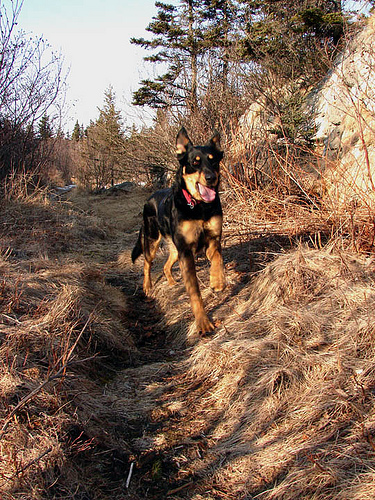}}\\
      \midrule
      
      \textbf{ref.} & \redbf{ein brauner und ein schwarzer Hund} laufen auf \\
      & einem Pfad im Wald. \\
      
      SMT & \redbf{ein braun und schwarzer} Hund läuft auf einem \\
      & Pfad im Wald. \\
      
      NMT & ein \redbf{brauner Hund steht} an einem \redbf{Sand Strand}. \\
      
      IMG$_\text{1W}$ & ein braun-schwarzer Hund läuft auf einem Pfad im Wald. \\
      
      IMG$_\text{2W}$ & ein braun-schwarzer Hund läuft im Wald auf einem Pfad. \\
      
      IMG$_\text{E}$ & ein braun-schwarzer Hund läuft im Wald auf einem Pfad. \\
      
      IMG$_\text{D}$ & ein braun-schwarzer Hund läuft im Wald auf einem Pfad. \\
      \midrule
      
      \multicolumn{2}{c}{\includegraphics[height=130pt]{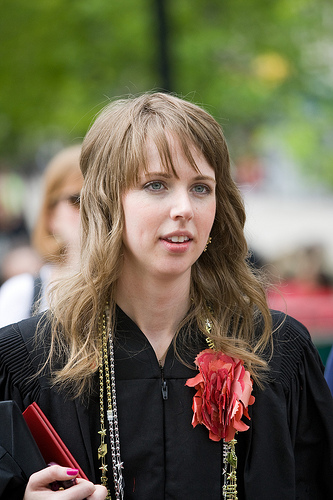}}\\
      \midrule
      
      \textbf{ref.} & eine Frau mit langen Haaren bei einer Abschluss Feier. \\
      
      SMT & eine Frau mit langen Haaren steht \redbf{an einem Abschluss} \\
      
      NMT & eine Frau mit langen Haaren \redbf{ist an einer StaZeremonie}. \\
      
      IMG$_\text{1W}$ & eine Frau mit langen Haaren \redbf{ist an einer warmen} \\
      & \redbf{Zeremonie teil}. \\
      
      IMG$_\text{2W}$ & eine Frau mit langen Haaren steht bei einer \bluebf{Hochzeit Feier}. \\
      
      IMG$_\text{E}$ & eine \bluebf{lang haarige} Frau bei einer \bluebf{olympischen} Zeremonie. \\
      
      IMG$_\text{D}$ & eine \bluebf{lang haarige} Frau bei einer \bluebf{olympischen} Zeremonie. \\
      
      
      
      
      
      
      
      
      
      \bottomrule
    \end{tabular}
  }
  \caption{Some translations for the M30k test set.}
  \label{tbl:translation-examples}
\end{table}

\section{Conclusions}
\label{sec:conclusion}

We have introduced different ideas to incorporate images into state-of-the-art attention-based NMT,
by using images as words in the source sentence, to initialise the encoder's hidden state and
as additional data in the initialisation of the decoder's hidden state.
We corroborate previous findings in that using image features directly at each time step of the decoder causes the model to overfit and prevents learning.
The intuition behind our effort is to use global image feature vectors to visually \emph{ground} translations and consequently increase translation quality.
Extensive experiments show that adding global image features into attention-based NMT is useful and improves over NMT and PBSMT as well as a strong multi-modal NMT baseline, according to all metrics evaluated.

In future work we will
conduct a more systematic study on the impact that synthetic back-translated data can have on multi-modal NMT,
and also investigate how to incorporate local, spatial-preserving image features.

\bibliography{phd}

\begin{thebibliography}{}
\expandafter\ifx\csname natexlab\endcsname\relax\def\natexlab#1{#1}\fi

\bibitem[{Bahdanau et~al.(2015)Bahdanau, Cho, and
  Bengio}]{BahdanauChoBengio2015}
Dzmitry Bahdanau, Kyunghyun Cho, and Yoshua Bengio. 2015.
\newblock {Neural Machine Translation by Jointly Learning to Align and
  Translate}.
\newblock In {\em International Conference on Learning Representations, ICLR
  2015\/}. San Diego, California.

\bibitem[{Bengio et~al.(2015)Bengio, Vinyals, Jaitly, and
  Shazeer}]{Bengioetal2015}
Samy Bengio, Oriol Vinyals, Navdeep Jaitly, and Noam~M. Shazeer. 2015.
\newblock \href{http://arxiv.org/abs/1506.03099}{{Scheduled Sampling for
  Sequence Prediction with Recurrent Neural Networks}}.
\newblock In {\em Advances in Neural Information Processing Systems, NIPS\/}.
\newblock
  \href{http://arxiv.org/abs/1506.03099}{http://arxiv.org/abs/1506.03099}.

\bibitem[{Bengio et~al.(2003)Bengio, Ducharme, Vincent, and
  Janvin}]{Bengioetal2003}
Yoshua Bengio, R{\'e}jean Ducharme, Pascal Vincent, and Christian Janvin. 2003.
\newblock \href{http://dl.acm.org/citation.cfm?id=944919.944966}{{A Neural
  Probabilistic Language Model}}.
\newblock {\em J. Mach. Learn. Res.\/} 3:1137--1155.
\newblock
  \href{http://dl.acm.org/citation.cfm?id=944919.944966}{http://dl.acm.org/citation.cfm?id=944919.944966}.

\bibitem[{Caglayan et~al.(2016)Caglayan, Aransa, Wang, Masana,
  Garc\'{i}a-Mart\'{i}nez, Bougares, Barrault, and van~de
  Weijer}]{Caglayanetal2016}
Ozan Caglayan, Walid Aransa, Yaxing Wang, Marc Masana, Mercedes
  Garc\'{i}a-Mart\'{i}nez, Fethi Bougares, Lo\"{i}c Barrault, and Joost van~de
  Weijer. 2016.
\newblock \href{http://www.aclweb.org/anthology/W/W16/W16-2358}{{Does
  Multimodality Help Human and Machine for Translation and Image Captioning?}}
\newblock In {\em Proceedings of the First Conference on Machine
  Translation\/}. Berlin, Germany, pages 627--633.
\newblock
  \href{http://www.aclweb.org/anthology/W/W16/W16-2358}{http://www.aclweb.org/anthology/W/W16/W16-2358}.

\bibitem[{Calixto et~al.(2012)Calixto, de~Campos, and Specia}]{Calixtoetal2012}
Iacer Calixto, Teofilo de~Campos, and Lucia Specia. 2012.
\newblock {Images as context in Statistical Machine Translation}.
\newblock In {\em Proceedings of the Workshop on Vision and Language, VL
  2012\/}. Sheffield, England.

\bibitem[{Calixto et~al.(2016)Calixto, Elliott, and
  Frank}]{CalixtoElliottFrank2016}
Iacer Calixto, Desmond Elliott, and Stella Frank. 2016.
\newblock \href{http://www.aclweb.org/anthology/W/W16/W16-2359}{{DCU-UvA
  Multimodal MT System Report}}.
\newblock In {\em Proceedings of the First Conference on Machine
  Translation\/}. Berlin, Germany, pages 634--638.
\newblock
  \href{http://www.aclweb.org/anthology/W/W16/W16-2359}{http://www.aclweb.org/anthology/W/W16/W16-2359}.

\bibitem[{Cho et~al.(2014{\natexlab{a}})Cho, van Merri{\"e}nboer, Bahdanau, and
  Bengio}]{Choetal2014b}
Kyunghyun Cho, Bart van Merri{\"e}nboer, Dzmitry Bahdanau, and Yoshua Bengio.
  2014{\natexlab{a}}.
\newblock On the properties of neural machine translation: Encoder--decoder
  approaches.
\newblock {\em Syntax, Semantics and Structure in Statistical Translation\/}
  page 103.

\bibitem[{Cho et~al.(2014{\natexlab{b}})Cho, van Merrienboer, Gulcehre,
  Bahdanau, Bougares, Schwenk, and Bengio}]{Choetal2014}
Kyunghyun Cho, Bart van Merrienboer, Caglar Gulcehre, Dzmitry Bahdanau, Fethi
  Bougares, Holger Schwenk, and Yoshua Bengio. 2014{\natexlab{b}}.
\newblock \href{http://www.aclweb.org/anthology/D14-1179}{{Learning Phrase
  Representations using RNN Encoder--Decoder for Statistical Machine
  Translation}}.
\newblock In {\em Proceedings of the 2014 Conference on Empirical Methods in
  Natural Language Processing (EMNLP)\/}. Doha, Qatar, pages 1724--1734.
\newblock
  \href{http://www.aclweb.org/anthology/D14-1179}{http://www.aclweb.org/anthology/D14-1179}.

\bibitem[{Clark et~al.(2011)Clark, Dyer, Lavie, and Smith}]{Clarketal2011}
Jonathan~H. Clark, Chris Dyer, Alon Lavie, and Noah~A. Smith. 2011.
\newblock \href{http://dl.acm.org/citation.cfm?id=2002736.2002774}{{Better
  Hypothesis Testing for Statistical Machine Translation: Controlling for
  Optimizer Instability}}.
\newblock In {\em Proceedings of the 49th Annual Meeting of the Association for
  Computational Linguistics: Human Language Technologies: Short Papers - Volume
  2\/}. Portland, Oregon, HLT '11, pages 176--181.
\newblock
  \href{http://dl.acm.org/citation.cfm?id=2002736.2002774}{http://dl.acm.org/citation.cfm?id=2002736.2002774}.

\bibitem[{Denkowski and Lavie(2014)}]{DenkowskiLavie2014}
Michael Denkowski and Alon Lavie. 2014.
\newblock {Meteor Universal: Language Specific Translation Evaluation for Any
  Target Language}.
\newblock In {\em Proceedings of the EACL 2014 Workshop on Statistical Machine
  Translation\/}.

\bibitem[{Donahue et~al.(2015)Donahue, Hendricks, Guadarrama, Rohrbach,
  Venugopalan, Darrell, and Saenko}]{Donahueetal2015}
Jeff Donahue, Lisa~Anne Hendricks, Sergio Guadarrama, Marcus Rohrbach,
  Subhashini Venugopalan, Trevor Darrell, and Kate Saenko. 2015.
\newblock {Long-term Recurrent Convolutional Networks for Visual Recognition
  and Description}.
\newblock In {\em Computer Vision and Pattern Recognition (CVPR), 2015 IEEE
  Conference on\/}. Boston, US, pages 2625--2634.

\bibitem[{Dong et~al.(2015)Dong, Wu, He, Yu, and Wang}]{Dongetal2015}
Daxiang Dong, Hua Wu, Wei He, Dianhai Yu, and Haifeng Wang. 2015.
\newblock \href{http://www.aclweb.org/anthology/P15-1166}{{Multi-Task Learning
  for Multiple Language Translation}}.
\newblock In {\em Proceedings of the 53rd Annual Meeting of the Association for
  Computational Linguistics and the 7th International Joint Conference on
  Natural Language Processing (Volume 1: Long Papers)\/}. Beijing, China, pages
  1723--1732.
\newblock
  \href{http://www.aclweb.org/anthology/P15-1166}{http://www.aclweb.org/anthology/P15-1166}.

\bibitem[{Elliott et~al.(2015)Elliott, Frank, and Hasler}]{Elliottetal2015}
Desmond Elliott, Stella Frank, and Eva Hasler. 2015.
\newblock \href{http://arxiv.org/abs/1510.04709}{Multi-language image
  description with neural sequence models}.
\newblock {\em CoRR\/} abs/1510.04709.
\newblock
  \href{http://arxiv.org/abs/1510.04709}{http://arxiv.org/abs/1510.04709}.

\bibitem[{Elliott et~al.(2016)Elliott, Frank, Sima'an, and
  Specia}]{ElliottFrankSimaanSpecia2016}
Desmond Elliott, Stella Frank, Khalil Sima'an, and Lucia Specia. 2016.
\newblock \href{http://aclweb.org/anthology/W/W16/W16-3210.pdf}{{Multi30K:
  Multilingual English-German Image Descriptions}}.
\newblock In {\em Proceedings of the 5th Workshop on Vision and Language,
  VL@ACL 2016\/}. Berlin, Germany.
\newblock
  \href{http://aclweb.org/anthology/W/W16/W16-3210.pdf}{http://aclweb.org/anthology/W/W16/W16-3210.pdf}.

\bibitem[{Firat et~al.(2016)Firat, Cho, and Bengio}]{Firatetal2016}
Orhan Firat, Kyunghyun Cho, and Yoshua Bengio. 2016.
\newblock \href{http://www.aclweb.org/anthology/N16-1101}{{Multi-Way,
  Multilingual Neural Machine Translation with a Shared Attention Mechanism}}.
\newblock In {\em Proceedings of the 2016 Conference of the North American
  Chapter of the Association for Computational Linguistics: Human Language
  Technologies\/}. San Diego, California, pages 866--875.
\newblock
  \href{http://www.aclweb.org/anthology/N16-1101}{http://www.aclweb.org/anthology/N16-1101}.

\bibitem[{Gal and Ghahramani(2016)}]{Gal2015}
Yarin Gal and Zoubin Ghahramani. 2016.
\newblock
  \href{http://papers.nips.cc/paper/6241-a-theoretically-grounded-application-of-dropout-in-recurrent-neural-networks.pdf}{{A
  Theoretically Grounded Application of Dropout in Recurrent Neural Networks}}.
\newblock In {\em Advances in Neural Information Processing Systems, NIPS\/},
  Barcelona, Spain, pages 1019--1027.
\newblock
  \href{http://papers.nips.cc/paper/6241-a-theoretically-grounded-application-of-dropout-in-recurrent-neural-networks.pdf}{http://papers.nips.cc/paper/6241-a-theoretically-grounded-application-of-dropout-in-recurrent-neural-networks.pdf}.

\bibitem[{Girshick et~al.(2014)Girshick, Donahue, Darrell, and
  Malik}]{Girshicketal2014}
Ross Girshick, Jeff Donahue, Trevor Darrell, and Jitendra Malik. 2014.
\newblock \href{https://doi.org/10.1109/CVPR.2014.81}{{Rich Feature Hierarchies
  for Accurate Object Detection and Semantic Segmentation}}.
\newblock In {\em Proceedings of the 2014 IEEE Conference on Computer Vision
  and Pattern Recognition\/}. Washington, DC, USA, CVPR '14, pages 580--587.
\newblock
  \href{https://doi.org/10.1109/CVPR.2014.81}{https://doi.org/10.1109/CVPR.2014.81}.

\bibitem[{Graves(2013)}]{Graves2013}
Alex Graves. 2013.
\newblock \href{http://arxiv.org/abs/1308.0850}{{Generating Sequences With
  Recurrent Neural Networks}}.
\newblock {\em CoRR\/} abs/1308.0850.
\newblock
  \href{http://arxiv.org/abs/1308.0850}{http://arxiv.org/abs/1308.0850}.

\bibitem[{Hitschler et~al.(2016)Hitschler, Schamoni, and
  Riezler}]{Hitschleretal2016}
Julian Hitschler, Shigehiko Schamoni, and Stefan Riezler. 2016.
\newblock \href{http://www.aclweb.org/anthology/P16-1227}{{Multimodal Pivots
  for Image Caption Translation}}.
\newblock In {\em Proceedings of the 54th Annual Meeting of the Association for
  Computational Linguistics (Volume 1: Long Papers)\/}. Berlin, Germany, pages
  2399--2409.
\newblock
  \href{http://www.aclweb.org/anthology/P16-1227}{http://www.aclweb.org/anthology/P16-1227}.

\bibitem[{Hochreiter and Schmidhuber(1997)}]{HochreiterSchmidhuber1997}
Sepp Hochreiter and J\"{u}rgen Schmidhuber. 1997.
\newblock \href{https://doi.org/10.1162/neco.1997.9.8.1735}{{Long Short-Term
  Memory}}.
\newblock {\em Neural Comput.\/} 9(8):1735--1780.
\newblock
  \href{https://doi.org/10.1162/neco.1997.9.8.1735}{https://doi.org/10.1162/neco.1997.9.8.1735}.

\bibitem[{Hodosh et~al.(2013)Hodosh, Young, and Hockenmaier}]{Hodoshetal2013}
Micah Hodosh, Peter Young, and Julia Hockenmaier. 2013.
\newblock \href{http://dl.acm.org/citation.cfm?id=2566972.2566993}{{Framing
  Image Description As a Ranking Task: Data, Models and Evaluation Metrics}}.
\newblock {\em J. Artif. Int. Res.\/} 47(1):853--899.
\newblock
  \href{http://dl.acm.org/citation.cfm?id=2566972.2566993}{http://dl.acm.org/citation.cfm?id=2566972.2566993}.

\bibitem[{Huang et~al.(2016)Huang, Liu, Shiang, Oh, and Dyer}]{Huangetal2016}
Po-Yao Huang, Frederick Liu, Sz-Rung Shiang, Jean Oh, and Chris Dyer. 2016.
\newblock
  \href{http://www.aclweb.org/anthology/W/W16/W16-2360}{{Attention-based
  Multimodal Neural Machine Translation}}.
\newblock In {\em Proceedings of the First Conference on Machine
  Translation\/}. Berlin, Germany, pages 639--645.
\newblock
  \href{http://www.aclweb.org/anthology/W/W16/W16-2360}{http://www.aclweb.org/anthology/W/W16/W16-2360}.

\bibitem[{Kalchbrenner and Blunsom(2013)}]{KalchbrennerBlunsom2013}
Nal Kalchbrenner and Phil Blunsom. 2013.
\newblock {Recurrent Continuous Translation Models}.
\newblock In {\em Proceedings of the 2013 Conference on Empirical Methods in
  Natural Language Processing, EMNLP 2013\/}. Seattle, pages 1700--1709.

\bibitem[{Karpathy and Fei-Fei(2015)}]{KarpathyFeiFei2015}
Andrej Karpathy and Li~Fei-Fei. 2015.
\newblock Deep visual-semantic alignments for generating image descriptions.
\newblock In {\em Proceedings of the IEEE Conference on Computer Vision and
  Pattern Recognition, CVPR 2015\/}. Boston, Massachusetts, pages 3128--3137.

\bibitem[{Kiros et~al.(2014)Kiros, Salakhutdinov, and Zemel}]{Kirosetal2014b}
Ryan Kiros, Ruslan Salakhutdinov, and Richard~S. Zemel. 2014.
\newblock \href{http://arxiv.org/abs/1411.2539}{Unifying visual-semantic
  embeddings with multimodal neural language models}.
\newblock {\em CoRR\/} abs/1411.2539.
\newblock
  \href{http://arxiv.org/abs/1411.2539}{http://arxiv.org/abs/1411.2539}.

\bibitem[{Kneser and Ney(1995)}]{KneserNey1995}
Reinhard Kneser and Hermann Ney. 1995.
\newblock Improved backing-off for m-gram language modeling.
\newblock In {\em In Proceedings of the IEEE International Conference on
  Acoustics, Speech and Signal Processing\/}. Detroit, Michigan, volume~I,
  pages 181--184.

\bibitem[{Koehn et~al.(2007)Koehn, Hoang, Birch, Callison-Burch, Federico,
  Bertoldi, Cowan, Shen, Moran, Zens, Dyer, Bojar, Constantin, and
  Herbst}]{Koehnetal2007}
Philipp Koehn, Hieu Hoang, Alexandra Birch, Chris Callison-Burch, Marcello
  Federico, Nicola Bertoldi, Brooke Cowan, Wade Shen, Christine Moran, Richard
  Zens, Chris Dyer, Ond\v{r}ej Bojar, Alexandra Constantin, and Evan Herbst.
  2007.
\newblock \href{http://dl.acm.org/citation.cfm?id=1557769.1557821}{{Moses: Open
  Source Toolkit for Statistical Machine Translation}}.
\newblock In {\em Proceedings of the 45th Annual Meeting of the ACL on
  Interactive Poster and Demonstration Sessions\/}. Association for
  Computational Linguistics, Prague, Czech Republic, ACL '07, pages 177--180.
\newblock
  \href{http://dl.acm.org/citation.cfm?id=1557769.1557821}{http://dl.acm.org/citation.cfm?id=1557769.1557821}.

\bibitem[{Libovick\'{y} et~al.(2016)Libovick\'{y}, Helcl, Tlust\'{y}, Bojar,
  and Pecina}]{Libovickyetal2016}
Jind\v{r}ich Libovick\'{y}, Jind\v{r}ich Helcl, Marek Tlust\'{y}, Ond\v{r}ej
  Bojar, and Pavel Pecina. 2016.
\newblock \href{http://www.aclweb.org/anthology/W/W16/W16-2361}{{CUNI System
  for WMT16 Automatic Post-Editing and Multimodal Translation Tasks}}.
\newblock In {\em Proceedings of the First Conference on Machine
  Translation\/}. Berlin, Germany, pages 646--654.
\newblock
  \href{http://www.aclweb.org/anthology/W/W16/W16-2361}{http://www.aclweb.org/anthology/W/W16/W16-2361}.

\bibitem[{Luong et~al.(2016)Luong, Le, Sutskever, Vinyals, and
  Kaiser}]{Luongetal2016}
Minh-Thang Luong, Quoc~V. Le, Ilya Sutskever, Oriol Vinyals, and Lukasz Kaiser.
  2016.
\newblock {Multi-Task Sequence to Sequence Learning}.
\newblock In {\em Proceedings of the International Conference on Learning
  Representations (ICLR), 2016\/}. San Juan, Puerto Rico.

\bibitem[{Luong et~al.(2015)Luong, Pham, and Manning}]{Luongetal2015}
Thang Luong, Hieu Pham, and Christopher~D. Manning. 2015.
\newblock {Effective Approaches to Attention-based Neural Machine Translation.}
\newblock In {\em Proceedings of the 2015 Conference on Empirical Methods in
  Natural Language Processing (EMNLP)\/}. Lisbon, Portugal, pages 1412--1421.

\bibitem[{Mao et~al.(2014)Mao, Xu, Yang, Wang, and Yuille}]{Maoetal2014}
Junhua Mao, Wei Xu, Yi~Yang, Jiang Wang, and Alan~L. Yuille. 2014.
\newblock \href{http://arxiv.org/abs/1410.1090}{{Explain Images with Multimodal
  Recurrent Neural Networks}}.
\newblock
  \href{http://arxiv.org/abs/1410.1090}{http://arxiv.org/abs/1410.1090}.

\bibitem[{Och(2003)}]{Och2003}
Franz~Josef Och. 2003.
\newblock \href{https://doi.org/10.3115/1075096.1075117}{{Minimum Error Rate
  Training in Statistical Machine Translation}}.
\newblock In {\em Proceedings of the 41st Annual Meeting on Association for
  Computational Linguistics - Volume 1\/}. Sapporo, Japan, ACL '03, pages
  160--167.
\newblock
  \href{https://doi.org/10.3115/1075096.1075117}{https://doi.org/10.3115/1075096.1075117}.

\bibitem[{Papineni et~al.(2002)Papineni, Roukos, Ward, and
  Zhu}]{Papinenietal2002}
Kishore Papineni, Salim Roukos, Todd Ward, and Wei-Jing Zhu. 2002.
\newblock \href{https://doi.org/10.3115/1073083.1073135}{{BLEU: A Method for
  Automatic Evaluation of Machine Translation}}.
\newblock In {\em Proceedings of the 40th Annual Meeting on Association for
  Computational Linguistics\/}. Philadelphia, Pennsylvania, ACL '02, pages
  311--318.
\newblock
  \href{https://doi.org/10.3115/1073083.1073135}{https://doi.org/10.3115/1073083.1073135}.

\bibitem[{Popovi\'{c}(2015)}]{Popovic2015}
Maja Popovi\'{c}. 2015.
\newblock \href{http://aclweb.org/anthology/W15-3049}{chrf: character n-gram
  f-score for automatic mt evaluation}.
\newblock In {\em Proceedings of the Tenth Workshop on Statistical Machine
  Translation\/}. Lisbon, Portugal, pages 392--395.
\newblock
  \href{http://aclweb.org/anthology/W15-3049}{http://aclweb.org/anthology/W15-3049}.

\bibitem[{Russakovsky et~al.(2015)Russakovsky, Deng, Su, Krause, Satheesh, Ma,
  Huang, Karpathy, Khosla, Bernstein, Berg, and Fei-Fei}]{Russakovskyetal2014}
Olga Russakovsky, Jia Deng, Hao Su, Jonathan Krause, Sanjeev Satheesh, Sean Ma,
  Zhiheng Huang, Andrej Karpathy, Aditya Khosla, Michael Bernstein,
  Alexander~C. Berg, and Li~Fei-Fei. 2015.
\newblock \href{https://doi.org/10.1007/s11263-015-0816-y}{{ImageNet Large
  Scale Visual Recognition Challenge}}.
\newblock {\em International Journal of Computer Vision (IJCV)\/}
  115(3):211--252.
\newblock
  \href{https://doi.org/10.1007/s11263-015-0816-y}{https://doi.org/10.1007/s11263-015-0816-y}.

\bibitem[{Sennrich et~al.(2016{\natexlab{a}})Sennrich, Haddow, and
  Birch}]{Sennrichetal2016a}
Rico Sennrich, Barry Haddow, and Alexandra Birch. 2016{\natexlab{a}}.
\newblock \href{http://www.aclweb.org/anthology/P16-1009}{{Improving Neural
  Machine Translation Models with Monolingual Data}}.
\newblock In {\em Proceedings of the 54th Annual Meeting of the Association for
  Computational Linguistics (Volume 1: Long Papers)\/}. Berlin, Germany, pages
  86--96.
\newblock
  \href{http://www.aclweb.org/anthology/P16-1009}{http://www.aclweb.org/anthology/P16-1009}.

\bibitem[{Sennrich et~al.(2016{\natexlab{b}})Sennrich, Haddow, and
  Birch}]{Sennrichetal2016}
Rico Sennrich, Barry Haddow, and Alexandra Birch. 2016{\natexlab{b}}.
\newblock \href{http://www.aclweb.org/anthology/P16-1162}{{Neural Machine
  Translation of Rare Words with Subword Units}}.
\newblock In {\em Proceedings of the 54th Annual Meeting of the Association for
  Computational Linguistics (Volume 1: Long Papers)\/}. Berlin, Germany, pages
  1715--1725.
\newblock
  \href{http://www.aclweb.org/anthology/P16-1162}{http://www.aclweb.org/anthology/P16-1162}.

\bibitem[{Shah et~al.(2016)Shah, Wang, and Specia}]{Shahetal2016}
Kashif Shah, Josiah Wang, and Lucia Specia. 2016.
\newblock
  \href{http://www.aclweb.org/anthology/W/W16/W16-2363}{{SHEF-Multimodal:
  Grounding Machine Translation on Images}}.
\newblock In {\em Proceedings of the First Conference on Machine
  Translation\/}. Berlin, Germany, pages 660--665.
\newblock
  \href{http://www.aclweb.org/anthology/W/W16/W16-2363}{http://www.aclweb.org/anthology/W/W16/W16-2363}.

\bibitem[{Simonyan and Zisserman(2014)}]{SimonyanZisserman2014}
K.~Simonyan and A.~Zisserman. 2014.
\newblock {Very Deep Convolutional Networks for Large-Scale Image Recognition}.
\newblock {\em CoRR\/} abs/1409.1556.

\bibitem[{Snover et~al.(2006)Snover, Dorr, Schwartz, Micciulla, and
  Makhoul}]{Snoveretal2006}
Matthew Snover, Bonnie Dorr, Richard Schwartz, Linnea Micciulla, and John
  Makhoul. 2006.
\newblock A study of translation edit rate with targeted human annotation.
\newblock In {\em In Proceedings of Association for Machine Translation in the
  Americas\/}. Cambridge, MA, pages 223--231.

\bibitem[{Specia et~al.(2016)Specia, Frank, Sima'an, and
  Elliott}]{Speciaetal2016}
Lucia Specia, Stella Frank, Khalil Sima'an, and Desmond Elliott. 2016.
\newblock \href{http://aclweb.org/anthology/W/W16/W16-2346.pdf}{{A Shared Task
  on Multimodal Machine Translation and Crosslingual Image Description}}.
\newblock In {\em Proceedings of the First Conference on Machine Translation,
  {WMT} 2016\/}. Berlin, Germany, pages 543--553.
\newblock
  \href{http://aclweb.org/anthology/W/W16/W16-2346.pdf}{http://aclweb.org/anthology/W/W16/W16-2346.pdf}.

\bibitem[{Stanojevi\'{c} et~al.(2015)Stanojevi\'{c}, Kamran, Koehn, and
  Bojar}]{Stanojevicetal2015}
Milo\v{s} Stanojevi\'{c}, Amir Kamran, Philipp Koehn, and Ond\v{r}ej Bojar.
  2015.
\newblock \href{http://aclweb.org/anthology/W15-3031}{{Results of the WMT15
  Metrics Shared Task}}.
\newblock In {\em Proceedings of the Tenth Workshop on Statistical Machine
  Translation\/}. Lisbon, Portugal, pages 256--273.
\newblock
  \href{http://aclweb.org/anthology/W15-3031}{http://aclweb.org/anthology/W15-3031}.

\bibitem[{Sutskever et~al.(2014)Sutskever, Vinyals, and
  Le}]{SutskeverVinyalsLe2014}
Ilya Sutskever, Oriol Vinyals, and Quoc~V Le. 2014.
\newblock {Sequence to Sequence Learning with Neural Networks}.
\newblock In {\em Advances in Neural Information Processing Systems\/}.
  Montr\'eal, Canada, pages 3104--3112.

\bibitem[{Tu et~al.(2016)Tu, Lu, Liu, Liu, and Li}]{Tuetal2016}
Zhaopeng Tu, Zhengdong Lu, Yang Liu, Xiaohua Liu, and Hang Li. 2016.
\newblock \href{http://www.aclweb.org/anthology/P16-1008}{{Modeling Coverage
  for Neural Machine Translation}}.
\newblock In {\em Proceedings of the 54th Annual Meeting of the Association for
  Computational Linguistics (Volume 1: Long Papers)\/}. Berlin, Germany, pages
  76--85.
\newblock
  \href{http://www.aclweb.org/anthology/P16-1008}{http://www.aclweb.org/anthology/P16-1008}.

\bibitem[{van Miltenburg(2015)}]{Miltenburg2015}
Emiel van Miltenburg. 2015.
\newblock Stereotyping and bias in the flickr30k dataset.
\newblock In {\em Proceedings of the Workshop on Multimodal Corpora,
  MMC-2016\/}. Portoro\v{z}, Slovenia, pages 1--4.

\bibitem[{Venugopalan et~al.(2015)Venugopalan, Rohrbach, Donahue, Mooney,
  Darrell, and Saenko}]{Venugopalanetal2015}
Subhashini Venugopalan, Marcus Rohrbach, Jeffrey Donahue, Raymond Mooney,
  Trevor Darrell, and Kate Saenko. 2015.
\newblock {Sequence to Sequence - Video to Text}.
\newblock In {\em Proceedings of the IEEE International Conference on Computer
  Vision\/}. Santiago, Chile, pages 4534--4542.

\bibitem[{Vinyals et~al.(2015)Vinyals, Toshev, Bengio, and
  Erhan}]{Vinyalsetal2014}
Oriol Vinyals, Alexander Toshev, Samy Bengio, and Dumitru Erhan. 2015.
\newblock Show and tell: {A} neural image caption generator.
\newblock In {\em {IEEE} Conference on Computer Vision and Pattern Recognition,
  {CVPR} 2015\/}. Boston, Massachusetts, pages 3156--3164.

\bibitem[{Xu et~al.(2015)Xu, Ba, Kiros, Cho, Courville, Salakhudinov, Zemel,
  and Bengio}]{Xuetal2015}
Kelvin Xu, Jimmy Ba, Ryan Kiros, Kyunghyun Cho, Aaron Courville, Ruslan
  Salakhudinov, Rich Zemel, and Yoshua Bengio. 2015.
\newblock \href{http://jmlr.org/proceedings/papers/v37/xuc15.pdf}{Show, attend
  and tell: Neural image caption generation with visual attention}.
\newblock In {\em Proceedings of the 32nd International Conference on Machine
  Learning (ICML-15)\/}. JMLR Workshop and Conference Proceedings, Lille,
  France, pages 2048--2057.
\newblock
  \href{http://jmlr.org/proceedings/papers/v37/xuc15.pdf}{http://jmlr.org/proceedings/papers/v37/xuc15.pdf}.

\bibitem[{Young et~al.(2014)Young, Lai, Hodosh, and
  Hockenmaier}]{Youngetal2014}
Peter Young, Alice Lai, Micah Hodosh, and Julia Hockenmaier. 2014.
\newblock {From image descriptions to visual denotations: New similarity
  metrics for semantic inference over event descriptions}.
\newblock {\em Transactions of the Association for Computational Linguistics\/}
  2:67--78.

\bibitem[{Zeiler(2012)}]{Zeiler2012}
Matthew~D. Zeiler. 2012.
\newblock \href{http://arxiv.org/abs/1212.5701}{{ADADELTA: An Adaptive Learning
  Rate Method}}.
\newblock {\em CoRR\/} abs/1212.5701.
\newblock
  \href{http://arxiv.org/abs/1212.5701}{http://arxiv.org/abs/1212.5701}.

\end{thebibliography}
\bibliographystyle{acl_natbib}

\end{document}